# ICD Codes are Insufficient to Create Datasets for Machine Learning: An Evaluation Using All of Us Data for Coccidioidomycosis and Myocardial Infarction

Abigail E. Whitlock
*Management and Information Systems*
*University of Arizona*
Tucson, United States
awhitlock@arizona.edu

Gondy Leroy, PhD
*Management and Information Systems*
*University of Arizona*
Tucson, United States
gondyleroy@arizona.edu

Fariba M. Donovan, MD, PhD
*Division of Infectious Disease*
*Valley Fever Center for Excellence*
Tucson, United States
faribadonovan@arizona.edu

John N. Galgiani, MD
*Division of Infectious Disease*
*Valley Fever Center for Excellence*
Tucson, United States
spherule@arizona.edu

*Abstract*—In medicine, machine learning (ML) datasets are often built using the International Classification of Diseases (ICD) codes. As new models are being developed, there is a need for larger datasets. However, ICD codes are intended for billing. We aim to determine how suitable ICD codes are for creating datasets to train ML models. We focused on a rare and common disease using the All of Us database. First, we compared the patient cohort created using ICD codes for Valley fever (coccidioidomycosis, CM) with that identified via serological confirmation. Second, we compared two similarly created patient cohorts for myocardial infarction (MI) patients. We identified significant discrepancies between these two groups, and the patient overlap was small. The CM cohort had 811 patients in the ICD-10 group, 619 patients in the positive-serology group, and 24 with both. The MI cohort had 14,875 patients in the ICD-10 group, 23,598 in the MI laboratory-confirmed group, and 6,531 in both. Demographics, rates of disease symptoms, and other clinical data varied across our case study cohorts.

*Keywords—ICD Codes, Valley fever, Myocardial Infarction*

## I. INTRODUCTION

Machine learning has been extensively used in medicine. With traditional machine learning (ML) algorithms, thousands of examples are needed to train models, e.g., 16,863 pediatric patients to assess the development of acute kidney injury [1], 790,470 examples to predict developing myelodysplastic syndrome [2], and clinical phenotyping using EHR data from 113,493 samples [3]. With increasingly sophisticated ML, larger data sets are needed. However, they are not always available; e.g., Betzler et al. used 6,066 retinal photographs [4], and Souza et al. used 1,880 MRI scans [5] with deep learning algorithms.

Larger datasets usually result in better outcomes and better representation. However, creating these datasets can be complicated, expensive, and time-consuming. A convenient approach often used to select EHR for datasets is the International Classification of Diseases (ICD) codes, which are used as identifiers and allow for the quick development of data sets. However, ICD codes are created for billing purposes and are not intended to be a final diagnostic label.

We evaluate the quality of ICD-based datasets for representing medical conditions. We use the All of Us database data for Valley fever and MI. Although the All-of-Us dataset has a higher minority representation than other available research data tools, we can still compare the differences between ICD codes and confirmatory diagnostic data.

## II. BACKGROUND

### A. International Classification of Diseases for Machine Learning Data Sets

The International Classification of Diseases (ICD) provides billable codes, and the Tenth Revision uses roughly 155,000 codes [6]. To comply with HIPAA, every healthcare provider must use ICD codes to classify and describe health conditions universally [7]. Sometimes, a physician may report a suspected condition via ICD code before testing it. If the patient does not have the suspected condition, the ICD code is not always corrected in the EHR. Additionally, clinically relevant ICD codes may not be used if they are not relevant for billing purposes [8].

In ML, ICD codes are often used as inclusion and exclusion criteria for a dataset. Table 1 shows a summary.

TABLE I. SUMMARY OF ICD LITERATURE

| Authors | Year | Number of Records | ICD Version | Other Characteristics for Dataset Creation |
|---|---|---|---|---|
| Dai et al. [9] | 2022 | 4,659 | 10 | None |
| You et al. [10] | 2023 | 475,611 | 9, 10 | Follow up and clinical criteria |
| Burns et al. [11] | 2022 | 4,295 | 10 | Clinical criteria |
| Pfaff et al. [12] | 2022 | 1,793,604 | 10 | Patients with clinical criteria and no ICD code included |
| Giannini et al. [13] | 2019 | 162,212 | 9 | Clinical criteria included |
| Fusar-Poli et al. [14] | 2019 | 91,199 | 10 | Time and location criteria included |
| Tran et al. [15] | 2022 | 1,380,740 | 10 | Patients with missing variables excluded |
| Chiu et al. [16] | 2022 | 83,227 | 9 | Patients with incomplete records excluded |
| Brettle et. Al [17] | 2021 | 81,659 | 10 | None |
| Sahoo et al. [18] | 2022 | 164,025 | 10 | None |
| McMaster et al. [19] | 2019 | 245 | 10 | Clinical review of ICD codes |

*B. All-of-Us*

The All of Us program started in 2015 with the intention of recruiting health data from at least one million Americans. Patients from every state in the US have participated, with California and Arizona providing the most participants thus far [20]. After data is uploaded from a participant, it is deidentified and stored. All of Us uses the Observational Medical Outcomes Partnership (OMOP) infrastructure to standardize data [21].

Researchers can register with the All of Us workbench and are given access to a broad spectrum of data. The All of Us workbench allows researchers to build a cohort or dataset using various variables that can serve as inclusion or exclusion criteria.

III. EXAMPLES USED IN OUR STUDY:

*A. Valley fever and Myocardial Infarction (MI)*

We present rare and common condition examples and compare the datasets created using ICD codes or confirmatory diagnostic data.

Valley fever, or Coccidioidomycosis (CM), is a fungal disease caused by inhaling Coccidioides spores [23] found in the soil in endemic regions of the United States, historically, Arizona and California. Valley fever can be challenging to diagnose. Most patients show no symptoms after exposure, while others report fatigue, cough, chest pain, night sweats, and rash. Noticeably, Valley fever symptoms are similar to those of Community-Acquired Pneumonia (CAP), which can lead to misdiagnosis or late diagnosis [24].

In contrast to Valley fever, myocardial infarction (MI) is a common cardiovascular disease. MI, commonly known as "heart attack," occurs when a portion of the heart loses blood supply and causes tissue damage [25]. Symptoms of MI include chest pain/tightness, shortness of breath, anxiety, and fatigue [25]. Noticeably, women are less likely to experience chest pain than men and tend to have symptoms of nausea and back or arm pain [25].

IV. METHODS

We compared datasets created using ICD codes and clinician-defined clinical criteria for inclusion. We used the All of Us Researcher workbench to build our datasets to search for symptoms and other clinical data (e.g., fever) or the ICD code (e.g., R50).

Patients are included in the ICD-10 Valley fever and myocardial infarction cohorts if they received a B38 or I21 code at any clinical encounter. Similarly, patients are included in the clinical criteria data set for their cohort if they were found to have at least one value within the range described below for any of the tests at any clinical encounter. Clinician experts decided on specific thresholds (See Tables 2 and 3). Patients could have both and then were part of the overlap between groups.

Datasets:

- *Valley fever ICD-10-Dataset*: ICD-10 B38 code, including any patient with the code listed in an EHR.
- *Valley fever Clinical Criteria Dataset*: laboratory serological tests used in clinical practice (See Table 2) and appropriate cutoff depending on the test when presented with numerical values (See Table 3).
- *Myocardial Infarction ICD-10-Dataset*: ICD-10 I21 code, including any patient with the code listed in an EHR.
- *Myocardial Infarction Clinical Criteria Dataset*: Troponin I greater than or equal to 0.2 and Troponin T greater than or equal to 0.1.

We compared demographics, symptoms, and clinical markers (lab tests that could be markers of disease if they contain abnormal values) in both groups because these are variables that ML algorithms would typically use. In addition, as an exploratory analysis, we added data on antibiotic usage. This would interest these studies as input to the machine learning models in answer to public health questions.

As defined by medical experts, the symptoms of Valley fever that we tracked included fatigue, cough, fever, chest pain, shortness of breath, headache, night sweats, muscle aches, joint pain, and rash (for the symptoms) and Procalcitonin, C-reactive protein, Erythrocyte Sedimentation Rate, Leukocytes, Hemoglobin, Eosinophil Count, and Albumin (for the clinical markers) [24]. We used the same symptoms as our Valley fever

This project was supported by the National Library of Medicine (NLM) of the National Institutes of Health under award number R25LM01422.

cohorts for MI because some symptoms are common in myocardial infarction.

TABLE II. VALLEY FEVER CLINICAL CRITERIA DATASET – SEROLOGICAL TESTS

| Presence of: |
|---|
| Coccidioides immitis Ab in serum |
| Coccidioides immitis Ag in isolate by Immune Diffusion |
| Coccidioides immitis IgM Ab in serum by Immune Diffusion |
| Coccidioides sp Ab in specimen by Immune Diffusion |
| Coccidioides immitis Ab in serum by Immune Diffusion |
| Coccidioides immitis IgG Ab in serum by Immune Diffusion |
| Coccidioides immitis IgM Ab in serum by Immunoassay |
| Coccidioides sp IgG Ab in serum |
| Coccidioides immitis IgG Ab in serum by Immunoassay |
| Coccidioides sp F Ab in serum by Immune Diffusion |
| Coccidioides sp IgM Ab in serum or plasma |
| Coccidioides immitis IgG Ab in serum |
| Coccidioides sp Ab in serum |
| Coccidioides sp TP Ab in serum by Immune Diffusion |
| Coccidioides immitis Ag in isolate |
| Coccidioides immitis IgM Ab in serum |
| **Titer of:** |
| Coccidioides immitis Ab in serum by Complement Fixation |
| Coccidioides immitis Ab in serum by Immune Diffusion |
| Coccidioides sp Ab in serum by Complement Fixation |

TABLE III. VALLEY FEVER CLINICAL CRITERIA DATASET – NUMERICAL VALUES

| Test | Value |
|---|---|
| Complement Fixation | ≥ 2 |
| Immunoassay | > 0 |
| Enzyme Immunoassay | > 0.8 |

## V. RESULTS

### A. Example 1: Valley fever

Table 4 shows the demographic data for the three cohorts: 811 cases using the ICD10 code, 619 cases using clinical tests, and 24 cases having both.

There are significant differences between the two groups. These differences include the percentage of Hispanic patients (23% for ICD-10, 50% for positive serology, and 42% for ICD-10 and positive serology) and the rate of fatigue (27% for ICD-10, 21% for positive serology, and 39% for ICD-10 and positive serology). Our ICD-10 cohort was predominantly white (63%), while our positive serology cohort and ICD-10 and positive serology cohort were largely unknown (54% and 46%, respectively).

The results for other variables are very similar. The median age in the ICD-10 cohort was 62, the median age in the positive serology cohort was 58, and the median age in the ICD-10 and positive serology cohort was 61. Our ICD-10 cohort and ICD-

TABLE IV. DEMOGRAPHICS FOR PATIENTS WITH VALLEY FEVER

| Demographics (N) | | | |
|---|---|---|---|
| | *ICD-10 Code: B38* | *Positive Serology for Valley fever* | *ICD-10 Code and Positive Serology Overlap* |
| Total | 811 | 619 | 24 |
| Median age, y (range) | 62 (20-96) | 58 (23-99) | 61 (25-92) |
| Sex, no. (%) | | | |
| Female | 423 (52.2) | 304 (49.1) | 12 (50.0) |
| Male | 359 (44.3) | 305 (49.3) | 11 (45.8) |
| Race, no. (%) | | | |
| African American | 80 (9.9) | 30 (4.8) | 4 (16.7) |
| Asian | 17 (2.1) | 35 (5.7) | 1 (4.2) |
| White | 507 (62.5) | 219 (35.4) | 8 (33.3) |
| Unknown | 207 (25.5) | 335 (54.1) | 11 (45.8) |
| Ethnicity, no. (%) | | | |
| Hispanic | 183 (22.6) | 308 (49.8) | 10 (41.7) |
| Non-Hispanic | 589 (72.6) | 291 (47.0) | 14 (58.3) |

10 and positive serology cohort had slightly more females than males (52% to 44%, 50% to 46%, respectively), while our positive serology cohort was equal at 49% each.

To compare the distributions of patients for the individual demographics in the three groups, we conducted a chi-square test of independence.

There was no significant association between sex and reporting Valley fever via ICD-10 code or clinical lab tests, $X^2$ (2, $N = 1414$) = 2.39, $p = 0.3$.

There was a significant association between race and reporting Valley fever via ICD-10 code or clinical lab tests, $X^2$ (6, N = 1454) = 154.40, $p < 0.001$. There was a significant association between ethnicity and reporting Valley fever via ICD-10 code or clinical lab tests, $X^2$ (2, N = 1395) = 112.91, $p < 0.001$.

Our clinical markers had wide ranges, but both cohorts had relatively similar median values. Our positive serology cohort had slightly elevated C-reactive protein (CRP), erythrocyte sedimentation rate (ESR), and Albumin. Leukocytes were more elevated in our ICD-10 cohort (See Table 5).

Figure 1 shows the percentage of patients with symptoms for each cohort in our Valley Fever case study. There was a statistically significant association between reporting fatigue ($X^2$ (2, $N = 1454$) = 6.75, $p = 0.03$), shortness of breath ($X^2$ (2, $N = 1454$) = 36.29, $p < 0.001$), night sweats ($X^2$ (2, $N = 1454$) = 13.56, $p = 0.001$), and joint pain ($X^2$ (2, $N = 1454$) = 8.00, $p = 0.02$) and reporting Valley fever via either ICD-10 code or laboratory confirmation.

The ICD-10 cohort had 27.4% of participants reporting fatigue, 47.3% reporting shortness of breath, 0.1% reporting night sweats, and 11.3% reporting joint pain. The positive serology cohort had 21.5% of participants reporting fatigue, 49.1% reporting shortness of breath, 0% reporting night sweats, and 13.9% reporting joint pain. The ICD-10 and positive serology cohort had 29.2% of participants reporting fatigue, 66.7% reporting shortness of breath, 0% reporting night sweats, and 29.2% reporting joint pain.

TABLE V. CLINICAL LAB TESTS FOR PATIENTS WITH VALLEY FEVER BETWEEN COHORTS

| Lab Tests (means) | | | |
|---|---|---|---|
| | B38 Code (range) | Positive Serology for Valley Fever (range) | ICD-10 Code and Positive Serology Overlap (range) |
| Procalcitonin, ng/mL | 0.375 (0.03-37.46) | 0.36 (0.02-136.5) | 1.34 (0.03-37.46) |
| C-reactive protein, mg/L | 0.4 (0-3) | 1.2 (0-41.1) | 1.95 (0.1-23.1) |
| ESR, mm/h | 33 (1-145) | 39 (0-140) | 23 (1.0-120.0) |
| Leukocytes, x 10^3 cells/mm^3 | 7.6 (0-12700) | 2 (0-182) | 2 (1.0-151.0) |
| Hemoglobin, g/dl | 10.4 (0-218) | 11.6 (0-244) | 13.5 (6.5-188.0) |
| Eosinophil count, x 10^3/microL | 0.3 (0-1066) | 0.1 (0-2175.0) | 0.1 (0.0-62.0) |
| Albumin, g/dL | 20 (0-10012) | 33 (0-3310) | 32 (0.2-61.0) |

All cohorts had similar rates in the following symptoms: cough, fever, chest pain, headache, muscle aches, and rash, and there were no statistically significant associations between cough ($X^2$ (2, $N$ = 1454) = 0.04, $p$ = 0.98), fever ($X^2$ (2, $N$ = 1454) = 0.60, $p$ = 0.74), chest pain ($X^2$ (2, $N$ = 1454) = 0.69, $p$ = 0.71), headache ($X^2$ (2, $N$ = 1454) = 1.60, $p$ = 0.45), muscle aches ($X^2$ (2, $N$ = 1454) = 3.04, $p$ = 0.22), and rash ($X^2$ (2, $N$ = 1454) = 2.70, $p$ = 0.26) and reporting Valley fever via either ICD-10 code or laboratory confirmation.

We also investigated antibiotic use in our two cohorts. In our ICD-10 cohort, 33% of patients were on an antibiotic within the month before their ICD-10 code designation of B38. Our positive serology cohort found that 40% of patients were on an antibiotic within the month before their serological confirmation of Valley fever. Our ICD-10 and positive serology cohort found that 38% of patients were on antibiotics within the month before their ICD-10 code and positive serology.

## B. Example 2: Myocardial Infarction

Table 6 shows the demographic data for the three cohorts: 14,875 cases using the ICD-10 code, 23,598 cases using clinical tests, and 6,531 cases had both.

The main differences were found for sex. Our ICD-10 cohort and ICD-10 with laboratory confirmation had more males than females (56% to 42%, 55% to 43%), while our laboratory confirmation cohort had more females than males (56% to 42%). Our chi-square analysis confirmed this difference to be significant, $X^2$ (2, N = 44021) = 796.63, $p$ <0.001. There was also a statistically significant association between race and reporting myocardial infarction via either ICD-10 code or laboratory confirmation, $X^2$ (6, N = 44612) =

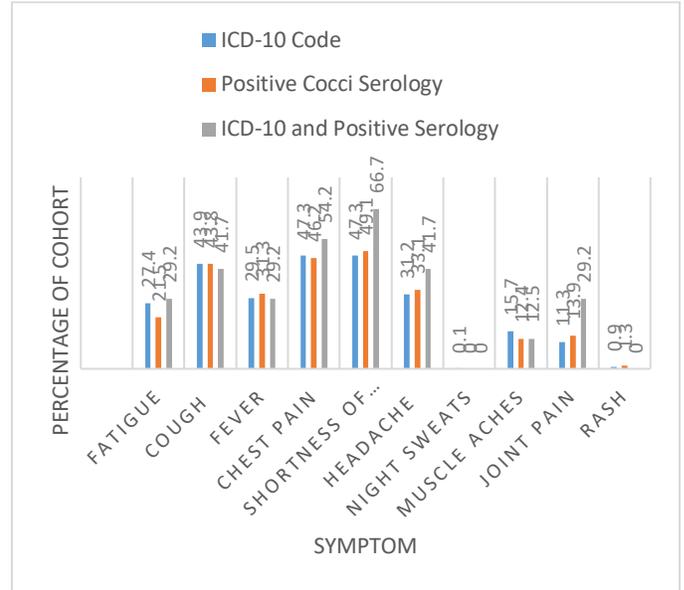

Fig. 1. Percentage of patients with Valley Fever and specific symptoms in ICD-10 and positive-serology cohorts.

TABLE VI. DEMOGRAPHICS FOR PATIENTS WITH MYOCARDIAL INFARCTION BETWEEN COHORTS

| Demographics (N) | | | |
|---|---|---|---|
| | ICD-10 Code I21 | Positive Serology for Myocardial Infarction | ICD-10 Code and Laboratory Confirmation Overlap |
| Total | 14,875 | 23,598 | 6,531 |
| Median age, y (range) | 68 (21-106) | 66 (20-106) | 69 (23-106) |
| Sex, no. (%) | | | |
| Female | 6,274 (42.2) | 13,121 (55.6) | 2,791 (42.7) |
| Male | 8,271 (55.6) | 9,981 (41.9) | 3,583 (54.9) |
| Race, no. (%) | | | |
| African American | 3,404 (22.9) | 5,199 (22.0) | 1,511 (23.1) |
| Asian | 200 (1.3) | 238 (1.0) | 89 (1.4) |
| White | 8,331 (56.0) | 12,443 (52.7) | 3,539 (52.4) |
| Unknown | 2,658 (17.9) | 5,608 (23.8) | 1,392 (21.3) |
| Ethnicity, no. (%) | | | |
| Hispanic | 2,198 (14.8) | 4,546 (19.3) | 1,080 (16.5) |
| Non-Hispanic | 12,027 (80.9) | 18,101 (76.7) | 5,183 (79.3) |

179.97, $p$ < 0.001. Ethnicity was also statistically significant, $X^2$ (2, N = 43135) = 129.63, $p$ < 0.001.

Other variables showed little difference: The median age in the ICD-10 cohort was 68, the median age in the laboratory confirmation cohort was 66, and the median age in the ICD-10 and laboratory confirmation cohort was 69.

Our clinical markers had wide ranges, but the median values were remarkably similar between cohorts. For example, procalcitonin, C-reactive protein, leukocytes, eosinophil count, and albumin were all within a two-point range of each other (see Table 7).

Figure 2 shows a comparison of the symptoms in our myocardial infarction groups. Both cohorts had similar rates of symptoms. There were statistically significant associations between fatigue ($X^2$ (2, $N$ = 45004) = 56.07, $p < 0.001$), cough ($X^2$ (2, $N$ = 45004) = 79.44, $p < 0.001$), fever ($X^2$ (2, $N$ = 45004) = 72.08, $p < 0.001$), chest pain ($X^2$ (2, $N$ = 45004) = 323.98, $p < 0.001$), shortness of breath ($X^2$ (2, $N$ = 45004) = 262.70, $p < 0.001$), headache ($X^2$ (2, $N$ = 45004) = 117.01, $p < 0.1$), muscle aches ($X^2$ (2, $N$ = 45004) = 19.09, $p < 0.001$), and joint pain ($X^2$ (2, $N$ = 45004) = 22.96, $p < 0.001$) and reporting myocardial infarction via either ICD-10 code or laboratory confirmation.

There was no statistically significant association for night sweats ($X^2$ (2, $N$ = 45004) = 0.73, $p = 0.69$) or rash ($X^2$ (2, $N$ = 45004) = 4.87, $p = 0.09$) and reporting myocardial infarction via either ICD-10 code or laboratory confirmation.

TABLE VII. CLINICAL LAB TESTS FOR PATIENTS WITH MYOCARDIAL INFARCTION BETWEEN COHORTS

| Lab Tests | | | |
|---|---|---|---|
| | *I21 Code* | *Positive Serology for Myocardial Infarction* | *ICD-10 Code and Laboratory Confirmation Overlap* |
| Procalcitonin, ng/mL | 0.21 (0-55.32) | 0.17 (0-471.38) | 0.22 (0.0-57.09) |
| C-reactive protein, mg/L | 4.65 (0-1900) | 3 (0-665) | 4.14 (0.0-603.7) |
| ESR, mm/h | 26 (0-499) | 24 (0-666) | 28.0 (0.0-499.0) |
| Leukocytes, x $10^3$ cells/mm$^3$ | 2 (0-16,865) | 3 (0-16865) | 3.0 (0.0-16865.0) |
| Hemoglobin, g/dL | 108 (0-1,420) | 109 (0-1480) | 106.0 (0.0-522.0) |
| Eosinophil count, x $10^3$/microL | 0.16 (0-7,400) | 0.11 (0-2850) | 0.13 (0.0-2850.0) |
| Albumin, g/dL | 36 (0-85,800) | 37 (0-85800) | 36.0 (0.0-85800.0) |

We investigated antibiotic use in our myocardial infarction cohorts as well. For our ICD-10 cohort, we found that 30.1% of our patients were on an antibiotic the month before their ICD-10 code of I21. For our laboratory confirmation cohort, we found that 10.3% of our patients were on an antibiotic the month before their laboratory confirmation for myocardial infarction. For our ICD-10 and laboratory confirmation, we found that 37% of our patients were on an antibiotic the month before their ICD-10 and laboratory confirmation.

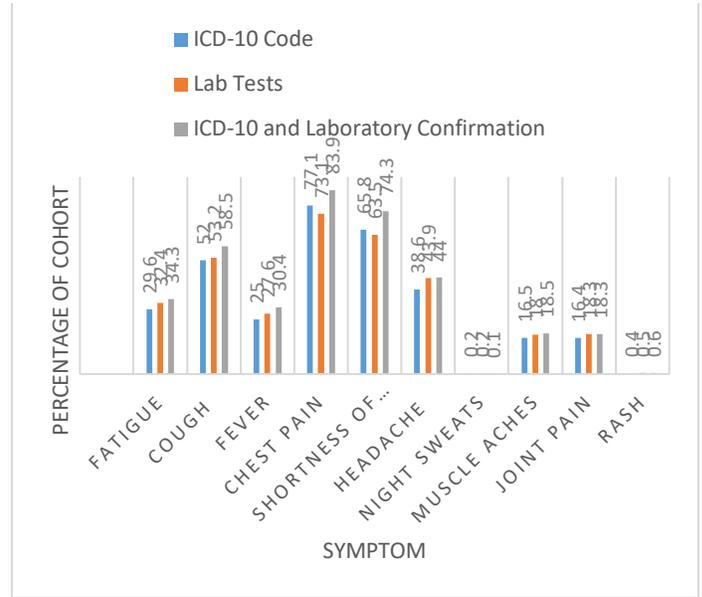

Fig. 2. Percentage of patients with MI and specific symptoms in ICD-10 and laboratory confirmation cohorts.

Notably, antibiotic usage is not usually clinically relevant to myocardial infarction unless the physician is also worried about pneumonia. It is used here as an example data point.

## VI. CONCLUSION

Underlying datasets are critical to the success of ML projects. We systematically compared ICD-10 codes, i.e., a commonly used approach to creating datasets, with a more clinically informed approach using serological tests using the All of Us database to provide an objective evaluation using the same underlying EHR for Valley fever and myocardial infarction. We analyzed demographics, clinical symptoms, and laboratory markers for each cohort. Overall, we found significant discrepancies between cohorts in both case studies. There was extremely little overlap between the ICD and clinically diagnosed groups, indicating that while symptoms are similar, relatively few patients have both an ICD code for a condition and the confirmatory symptoms and clinical tests one would expect. This is an essential distinction if ICD codes are used to build a machine-learning model based on the patient's clinical data.

The discrepancy in cohorts reflects the different intentions of ICD codes versus diagnostic test values. These are very distinct subgroups of patients. This is because ICD-10 codes are used when, e.g., ordering tests but not adjusted when the results return, and vice versa, ICD-10 codes are not needed when something else is billed. Additionally, cohorts may have incorrect data due to inaccurate reporting or coding. Furthermore, while we expect these cohorts to represent distinct populations, they may represent essentially the same group of patients, which could explain why we do not see the expected results.

The different numbers of patients in our cohorts for both case studies demonstrate that knowledge of the studied disease is critical for building a database. ML projects using ICD codes to predict diagnosis are likely predicting billing labels, not diagnosis. This difference is essential to know and consider when using ML in clinical care. With the increasing use of ML in practice, evaluating model performance will be insufficient, and more attention will be paid to data set creation.

While the All of Us data was an excellent tool for our comparison, more information may be needed to study the conditions. While All of Us can be utilized to search for patients by symptoms, access to the free text portion of an EHR is essential for accurate and more detailed information.

Finally, we acknowledge that our study has limitations. Cohorts were created based on the presence of an ICD code or serology tests without considering the timeline. Also, we only analyzed ICD-10 codes; further research must be done to analyze ICD-9 and ICD-11 codes. Additionally, we only evaluated two conditions: one more common, definitively diagnosed, and found across the United States (MI), and one less common, frequently misdiagnosed, and endemic to only California, Arizona, and Washington (Valley fever).

ACKNOWLEDGMENTS

Part of this project was supported by the National Library of Medicine (NLM) of the National Institutes of Health under award number R25LM01422 and an All of Us Researcher Engagement Award from the Office of the Director under award OT2OD036485.

This study used data from the *All of Us* Research Program's Controlled Tier Dataset v7, available to all authorized users on the Researcher Workbench. The *All of Us* Research Program is supported by the National Institutes of Health, Office of the Director: Regional Medical Centers: 1 OT2 OD026549; 1 OT2 OD026554; 1 OT2 OD026557; 1 OT2 OD026556; 1 OT2 OD026550; 1 OT2 OD 026552; 1 OT2 OD026553; 1 OT2 OD026548; 1 OT2 OD026551; 1 OT2 OD026555; IAA #: AOD 16037; Federally Qualified Health Centers: HHSN 263201600085U; Data and Research Center: 5 U2C OD023196; Biobank: 1 U24 OD023121; The Participant Center: U24 OD023176; Participant Technology Systems Center: 1 U24 OD023163; Communications and Engagement: 3 OT2 OD023205; 3 OT2 OD023206; and Community Partners: 1 OT2 OD025277; 3 OT2 OD025315; 1 OT2 OD025337; 1 OT2 OD025276. In addition, the All of Us Research Program would not be possible without the partnership of its participants.